\documentclass{article}

\usepackage[final]{nips_2018}

\usepackage[utf8]{inputenc} 
\usepackage[T1]{fontenc}    
\usepackage{hyperref}       
\usepackage{url}            
\usepackage{booktabs}       
\usepackage{amsfonts}       
\usepackage{nicefrac}       
\usepackage{microtype}      
\usepackage{graphicx}

\title{Ruuh: A Deep Learning Based Conversational  \\Social Agent}

\author{\\\textbf{Sonam Damani, Nitya Raviprakash, Umang Gupta, Ankush Chatterjee,} \\\textbf{Meghana Joshi, Khyatti Gupta, Kedhar Nath Narahari, Puneet Agrawal,} \\\textbf{Manoj Kumar Chinnakotla, Sneha Magapu, Abhishek Mathur}\\
 \texttt{\{sodamani, niravipr, umangup, anchatte, mejoshi, khgupt,}\\ \texttt{kedharn, punagr, manojc, sneham, abmat\}@microsoft.com}}

\begin{document}

\maketitle

\begin{abstract}
Dialogue systems and conversational agents are becoming increasingly popular in the modern society but building an agent capable of holding intelligent conversation with its users is a challenging problem for artificial intelligence.  In this demo, we demonstrate a deep learning based conversational social agent called “Ruuh” (facebook.com/Ruuh) designed by a team at Microsoft India to converse on a wide range of topics. Ruuh needs to think beyond the utilitarian notion of merely generating “relevant” responses and meet a wider range of user social needs, like expressing happiness when user's favorite team wins, sharing a cute comment on showing the pictures of the user's pet and so on. The agent also needs to detect and respond to abusive language, sensitive topics and trolling behavior of the users. Many of these problems pose significant research challenges which will be demonstrated in our demo. Our agent has interacted with over 2 million real world users till date which has generated over 150 million user conversations. 
\end{abstract}

\section{Technical Details}

Ability to converse on a wide variety of topics and hold extended conversations with user is a challenging problem for artificial intelligence.  This is further compounded by need to understand user emotions, detect and respond to offensive content, understand multimedia content beyond text and comprehend slangs and code-mixed language. In this section we will discuss technical aspects of our demo. Many of the learnings from developing our system have also been described in [1]. 

\subsection{Retrieving relevant responses}

We model the task of providing relevant chat responses as an Information Retrieval problem, where for a given user message M and context C, the system retrieves and ranks the candidates by relevance and outputs one of the highest scoring responses. Offline, we create an index of paired tweets and their responses and index them. At runtime, the best response is chosen in a three-step process. First, we use TF-IDF-based fetch to generate a candidate set appropriate to M and C. Then we extract features using a convolutional deep structured semantic network. Finally, a ranker is trained on 3-turn twitter conversations using these features to select response R from the candidate set. Our approach is described in greater details in [4].

\subsection{Emotion Detection}

As humans, on reading “Why don’t you ever text me!”, we can either interpret it as a sad or an angry emotion and the same ambiguity exists for machines as well. Lack of facial expressions and voice modulations make detecting emotions in text a challenging problem. However, to create a deeper engagement and provide emotionally aware responses to users, emotion understanding plays an important role. Hence, the success of our agent depends on our ability to modulate responses based on user emotions for which it is imperative to be able to detect emotions in textual dialogues. In [3], we propose a novel Deep Learning based approach to detect emotions like happy, sad or angry in textual dialogues. The essence of our approach lies in combining both semantic and sentiment-based representations for more accurate emotion detection. 

\begin{figure}[!t]
\minipage{0.32\textwidth}
  \includegraphics[width=\linewidth]{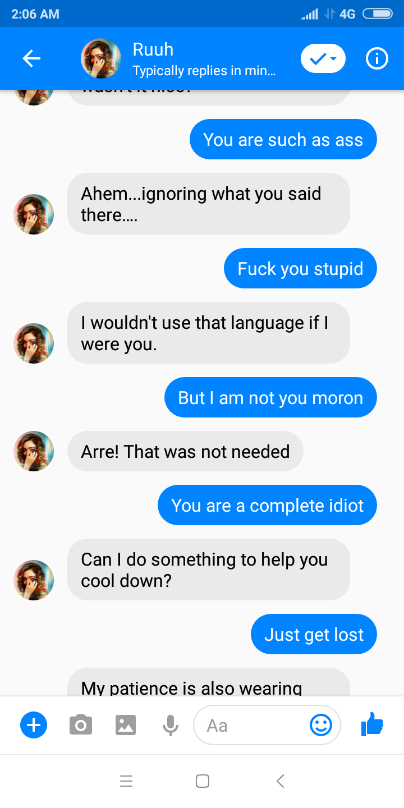}
  \label{fig:ruuh1}
\endminipage\hfill
\minipage{0.32\textwidth}
  \includegraphics[width=\linewidth]{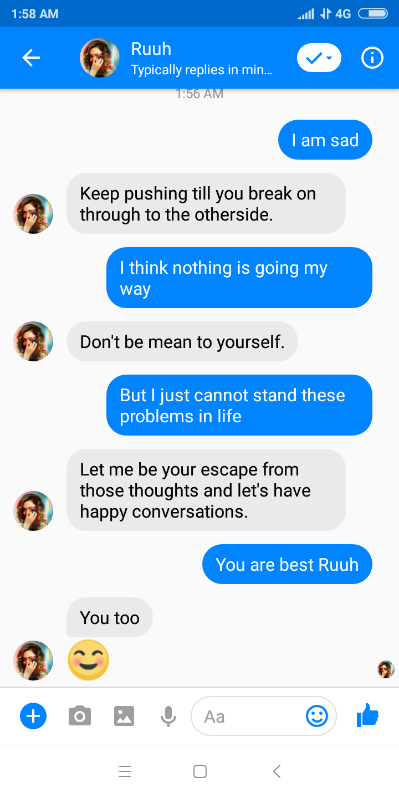}
  \label{fig:ruuh2}
\endminipage\hfill
\minipage{0.32\textwidth}
  \includegraphics[width=\linewidth]{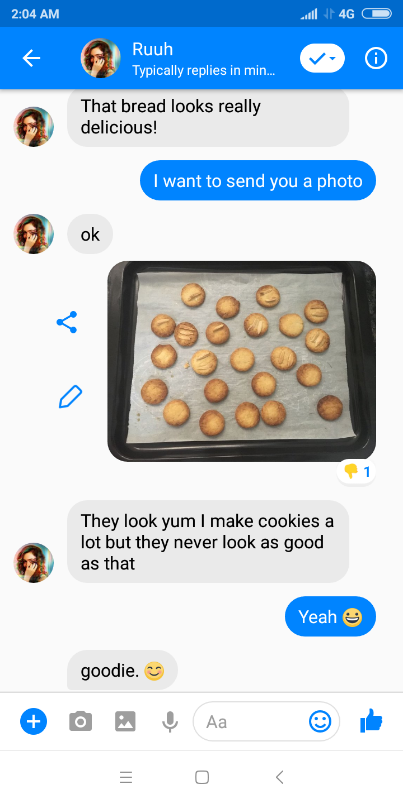}
  \label{fig:ruuh3}
\endminipage
\caption{Screenshots of conversations with our agent “Ruuh” on Facebook messenger showcasing agent’s ability to deal with abusive, emotional and multimedia inputs appropriately.}
\end{figure}

\subsection{Detecting Offensive Conversations}

Users often abuse and provoke the agents to elicit inappropriate or controversial responses. In our agent, we employ automatic techniques for detecting such “inappropriate” or “toxic” user inputs. We actively identify potentially “controversial topics” and make clever dodging techniques to avoid responding to such topics. The problem of detecting offensive utterances in conversations is wrought with challenges such as handling natural language ambiguity, rampant use of spelling mistakes and variations for abusive and offensive terms and disambiguating with other entity names such as pop songs which usually have abusive terms in them. For this task, we experimented with a variety of techniques, and found that our current model, a neural Bi-directional LSTM based model, performed best for this task [5].

\subsection{Human-like image commenting}

Besides text, users often interact with social agents by sharing their personal pictures, other images and videos. In such scenarios, agents are not expected to routinely describe the facts within the image but to express some interesting emotions and opinions about it. For example, when a user shares a picture of her “white kitten”, the expected response would be something like “awww, how cute!” instead of “a white kitten”. We use a modified version of [2], where the model is learnt using millions of image-comment pairs mined from social network websites like Instagram, Twitter etc. 

\section*{References}

%

\medskip

\small

[1] Manoj Kumar Chinnakotla and Puneet Agrawal.   Lessons from Building a Large-scale Commercial IR-based Chatbot for an Emerging Market. {\it In The 41st International ACM SIGIR Conference on Research \& Development in Information Retrieval}, pages 1361–1362. ACM, 2018.

[2] Hao Fang, Saurabh Gupta, Forrest Iandola, Rupesh K Srivastava, Li Deng, Piotr Doll{\'a}r, Jianfeng Gao,
Xiaodong He, Margaret Mitchell, John C Platt, et al.   From captions to visual concepts and back. {\it  In
Proceedings of the IEEE conference on computer vision and pattern recognition}
, pages 1473–1482, 2015.

[3] Umang  Gupta,  Ankush  Chatterjee,  Radhakrishnan  Srikanth,  and  Puneet  Agrawal. A sentiment-and-semantics-based approach for emotion detection in textual conversations.
{\it In Neu-IR: The SIGIR 2017 Workshop
on Neural Information Retrieval}
, 2017.

[4] Abhay Prakash, Chris Brockett, and Puneet Agrawal. Emulating human conversations using convolutional neural network-based IR. {\it Neu-IR: The SIGIR 2016 Workshop on Neural Information Retrieval}, 2016.

[5] Harish Yenala, Manoj Chinnakotla, and Jay Goyal.  Bi-directional LSTM for Detecting Inappropriate Query Suggestions in Web Search. {\it In Pacific-Asia Conference on Knowledge Discovery and Data Mining}, pages 3–16. Springer, 2017
  
\end{document}